\documentclass[runningheads]{llncs}

\usepackage{eccv}

\usepackage{eccvabbrv}

\usepackage{graphicx}
\usepackage{booktabs}

\usepackage{adjustbox} 
\usepackage{here} 
\usepackage{colortbl} 
\definecolor{MyLightGray}{gray}{0.97} 
\usepackage{hhline}  
\usepackage{multirow} 
\usepackage{caption}  
\usepackage{makecell}  

\usepackage[accsupp]{axessibility}  

\usepackage{hyperref}

\usepackage{orcidlink}

\begin{document}

\title{Mitigating Modality and Language-Style Gaps \\for Zero-Shot Video Moment Retrieval}

\titlerunning{Mitigating Modality and Language-Style Gaps for ZMR}

\newcommand{\samethanks}{\protect\footnotemark[\value{footnote}]}
\author{
Jihyun Lee\thanks{These authors contributed equally.}\orcidlink{0009-0008-6684-1845} \and
Cheol-Ho Cho\samethanks\orcidlink{0009-0003-6131-1136} \and
Woojin Jun\orcidlink{0009-0000-3912-4535} \and
Woojin Jeong\orcidlink{0009-0006-1751-7569} \and
Jae-Pil Heo\thanks{Corresponding author}\orcidlink{0000-0001-9684-7641}
}

\authorrunning{J.~Lee et al.}

\institute{
Sungkyunkwan University, Suwon, Republic of Korea
}




\maketitle

\begin{abstract}

Zero-shot video moment retrieval~(ZMR) aims to overcome the limitations of traditional approaches that require large-scale datasets annotated with text and its relevant temporal spans.
Despite advances in pre-trained vision–language models~(VLMs) and multimodal large language models~(MLLMs), existing ZMR methods still heavily depend on query-to-video content similarity, making them vulnerable to modality and language-style gaps. 
These gaps lead to unreliable span proposals and unstable moment retrieval results. 
To address this issue, we propose Self-Similarity-based Moment proposal and Scoring~(Self-SiMS) that instead exploits intrinsic relationships within videos, enabling robust span generation and scoring. 
By deriving self-similarity only from the video content, we circumvent the noisy and mismatched patterns of query–frame or query–caption similarities, thereby mitigating both modality and language-style gaps.
Furthermore, we introduce a query-aware MLLM-based reasoning stage to further sharpen alignment between text and video.
Extensive experiments demonstrate that Self-SiMS achieves the state-of-the-art performance across ZMR benchmarks.

\end{abstract}    
\begin{figure}[t]
    \begin{center}
        \includegraphics[width=\textwidth]{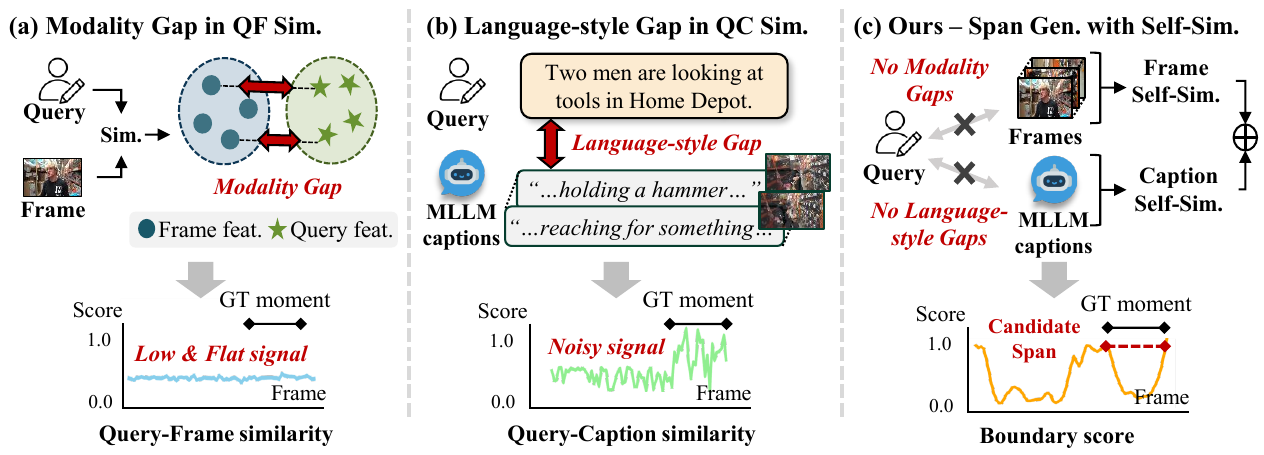}
    \end{center}
    \vspace{-0.5cm}
    \caption{Comparison of different similarity computation for candidate span generation.
(a) Query–Frame similarity suffers from a modality gap between textual and visual features, producing low and flat similarity signals.
(b) Query–Caption similarity alleviates modality issues but introduces a language-style gap, yielding noisy and inconsistent similarity patterns.
(c) Our method leverages intra-video self-similarity, avoiding such modality and language-style gaps, and produces more stable boundary scores.}
    \vspace{-0.6cm}
    \label{fig:motivation1}
\end{figure}

\section{Introduction}
\label{sec:intro}

With the rapid expansion of multimedia content, video has become a primary medium for obtaining information.
Consequently, retrieving specific moments from videos relevant to natural language queries has garnered significant attention across both academia and industry. 
A central task in this area is video moment retrieval (VMR)~\cite{barrett2015saying,gao2017tall,lei2021detecting,xu2024mh}, which aims to localize the most relevant temporal segment in a video based on a given text query. 
This task remains particularly challenging due to the need to pinpoint precise moments within videos that often contain multiple semantically similar scenes.

Previous studies have primarily focused on training VMR models using datasets annotated with temporal spans. 
However, such approaches require extensive manual labeling, which is both costly and labor-intensive. 
To overcome this limitation, recent research has explored zero-shot video moment retrieval (ZMR)~\cite{radford2021learning,nam2021zero,wang2022prompt,TFVTG,momentGPT}, leveraging vision-language models (VLMs)~\cite{li2022blip,li2023blip} and multimodal large language models (MLLMs)~\cite{touvron2023llama,touvron2023llama2,grattafiori2024llama3} pretrained on large-scale data to perform retrieval without additional supervision.
Although these methods have shown promising performance in ZMR, they still face a critical limitation stemming from the unreliable query-based similarity score used to generate and score candidate spans~(i.e., moment proposals).
Specifically, existing methods~\cite{TFVTG,momentGPT} rely on the similarity between text query and video contents, represented by either visual features or MLLM-generated captions.
However, these similarities are inherently affected by modality and language-style gaps, leading to the generation of unstable spans.
Fig.~\ref{fig:motivation1} conceptually summarizes how each approach computes similarity scores and visualizes the score tendencies observed at the video instance level when applying each method.

As illustrated in Fig.~\ref{fig:motivation1}(a), methods based on query--frame similarity~\cite{TFVTG} are affected by the modality gap between textual and visual representations. Even inside the ground-truth moment, the similarity can remain low and flat, making it difficult to distinguish the relevant temporal region from the surrounding context. 
Query--caption similarity~\cite{momentGPT} partially alleviates this issue by converting visual content into text, as shown in Fig.~\ref{fig:motivation1}(b). However, it introduces a language-style gap between human-written queries and MLLM-generated captions. Even for visually similar frames, the captions may focus on different aspects of the same scene, such as objects, actions, or fine-grained sub-events. For example, given the query "Two men are looking at tools in Home Depot," captions may alternately emphasize "holding a hammer" or "reaching for something." These descriptions can be valid, but their inconsistent focus leads to noisy and unstable query--caption similarity trajectories. \\
To avoid relying on such query-dependent signals, we use query-independent intra-video self-similarity for span generation. As shown in Fig.~\ref{fig:motivation1}(c), we compute self-similarity maps from both frame features and MLLM-generated caption features, and derive boundary scores by detecting changes in their temporal similarity patterns. These boundary scores are then used to generate candidate spans. Since this process depends only on the internal structure of the video, rather than query--frame or query--caption matching, the resulting spans are less affected by modality and language-style gaps.

\begin{figure}[t]
    \begin{center}
        \includegraphics[width=0.9\textwidth]{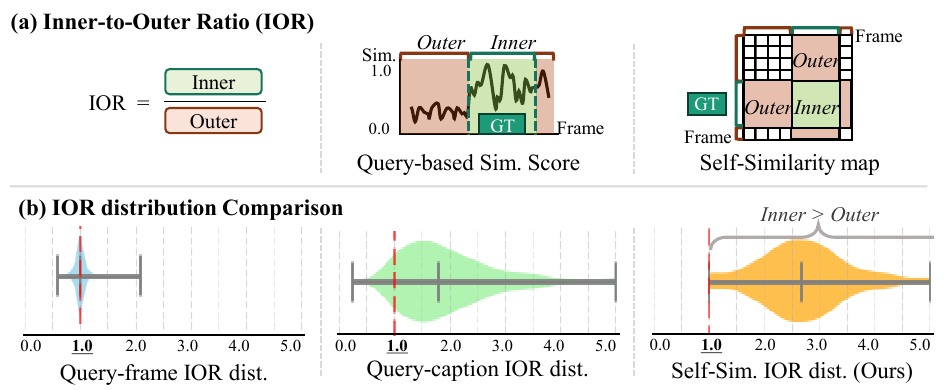}
    \end{center}
    \vspace{-0.6cm}
    \caption{Illustration and comparison of the Inner-to-Outer Ratio (IOR).
(a) The IOR measures how much higher the similarity is within the ground-truth (Inner) region than in the surrounding (Outer) area.
The middle and right examples illustrate how the Inner and Outer regions are obtained from query-based similarity scores and from the self-similarity map, respectively.
(b) Comparison of IOR distributions on the QVHighlights validation set shows that our method achieves consistently higher IOR values than query–frame and query–caption similarities, indicating stronger discrimination between relevant and irrelevant segments.}
    \vspace{-0.6cm}
    \label{fig:motivation2}
\end{figure}
While these trends are discovered at the instance level, similar patterns emerge at the dataset level as well.
To further examine these trends at the dataset level, we define the Inner-to-Outer Ratio (IOR) as a quantitative measure.
\textcolor{black}{Fig.~\ref{fig:motivation2}(a) illustrates the definition of IOR, which measures how much higher the similarity is within the ground-truth (Inner) region than in the surrounding (Outer) area.
An IOR above 1.0 indicates stronger similarity within the ground-truth span, whereas lower values imply weak discrimination, thereby measuring how well each method separates relevant from irrelevant segments.
Fig.~\ref{fig:motivation2}(b) reports the IOR distributions on the QVHighlights validation set as a representative example.
The query–frame similarity exhibits IOR values concentrated around 1.0, corresponding to the flat similarity patterns observed earlier and indicating that the modality gap leads to weak discrimination between inner and outer regions.
The query–caption similarity shows a broad and unstable distribution, reflecting the noisy similarity trajectories caused by inconsistent alignment between queries and generated captions.
In contrast, our method yields IOR values consistently above 1.0, indicating that it effectively captures intrinsic video structure and produces discriminative similarity scores across the dataset.
}

Building on this idea, we propose Self-Similarity-based Moment Proposal and Scoring (Self-SiMS), which leverages intra-video self-similarity for both candidate generation and span scoring. 
For span generation, self-similarity provides query-independent temporal structure, avoiding unstable query-frame or query-caption signals. 
For span scoring, we first use the most reliable query-caption signals within each span and then refine the score with self-similarity to reflect internal content consistency. This design reduces the effect of noisy query-caption alignment while preserving temporally coherent candidate spans.

Finally, to accurately select the final span from semantically similar span candidates, we introduce a query-aware MLLM-based re-ranking step. In this stage, representative frames from each span are directly evaluated against the query using an MLLM with a yes/no prompt. 
Instead of computing similarity after independently generating representations for the query and video, we leverage the MLLM's query-aware capabilities to further reduce both modality and language-style gaps, resulting in more accurate span scoring.

To sum up, our contributions are as follows:\textbf{}

\begin{itemize}
    \item We first explicitly identify and address the modality and language-style gaps in zero-shot video moment retrieval~(ZMR), where mismatches between text queries and video contents, whether visual features or MLLM-generated captions, can lead to inaccurate span scoring.
    \textcolor{black}{\item To mitigate these gaps, we propose Self-Similarity-based Moment proposal and Scoring~(Self-SiMS). 
    To the best of our knowledge, our work is the first to leverage temporal self-similarity for generating candidate spans in a training-free ZMR, and to score these spans using intra-video relations.}
    
    \item We further introduce a query-aware MLLM re-ranking step, where candidate spans are directly evaluated against the text query using an MLLM to mitigate the gaps above.
    
    \item Our method achieves the state-of-the-art performance on ZMR benchmarks, demonstrating the effectiveness of combining Self-SiMS with query-aware MLLM re-ranking in mitigating modality and language-style gaps and improving ZMR.
\end{itemize}
\section{Related works}

\noindent\textbf{Video moment retrieval.}
Video moment retrieval (VMR) aims to retrieve relevant temporal segments from an untrimmed video given a natural language query. 
Early work treated this as a query-to-segment matching problem using joint video–text feature learning and alignment techniques~\cite{barrett2015saying,gao2017tall,krishna2017dense}. 
With the introduction of large-scale datasets such as ActivityNet~\cite{caba2015activitynet} and QVHighlights~\cite{lei2021detecting}, subsequent methods explored attention mechanisms, cross-modal transformers, and contrastive learning for improved localization~\cite{lei2021detecting,sun2023gptsee,xu2024mh,xu2024multi,li2022blip,ma2023llavilo}. 
Recent approaches also leverage pre-trained vision–language models (VLMs) such as BLIP~\cite{li2022blip}, BLIP-2~\cite{li2023blip}, and instruction-tuned video models~\cite{zhang2023video,maaz2023video}, achieving strong results in moment retrieval and highlight detection. 
However, these advances still rely on large-scale span annotations, motivating research into weakly-supervised and zero-shot retrieval. \\

\noindent\textbf{Zero-shot video moment retrieval.}
Traditional VMR approaches rely on densely annotated datasets that precisely align queries with video moments~\cite{lei2021detecting,sun2023gptsee}, but such annotations are costly and labor-intensive, motivating research into zero-shot video moment retrieval (ZMR)~\cite{nam2021zero,wang2022prompt}. 
Early zero-shot methods leveraged pretrained VLMs such as CLIP~\cite{radford2021learning} to compute query–frame similarity without training data~\cite{nam2021zero}, yet they struggled to capture temporal relationships.
Moreover, recent studies show that off-the-shelf multimodal large language models~(MLLMs) can achieve competitive zero-shot performance without additional training~\cite{diwan2023zero,luo2024zero,wattasseril2023zero,huang2024vtimellm}. 
Despite these advances, existing methods still rely on query–video content similarity using either visual features or MLLM-generated captions for candidate span generation and scoring, which makes them vulnerable to modality and language-style gaps and limits precise moment localization. \\

\noindent\textbf{Modality and language-style gaps in text-visual alignment.}
Pretrained VLMs~\cite{radford2021learning, li2022blip, li2023blip} trained on large-scale text–image pairs have advanced cross-modal understanding, yet a modality gap remains, causing inaccurate similarity matching.
Prior works address this by introducing learnable gap parameters~\cite{xiao2025contrastive}, enhancing semantic modeling~\cite{liu2024towards}, or generating captions with MLLMs to bridge text and visuals.
However, MLLM-generated captions inherently create a language-style gap with human-written queries, which some studies mitigate by augmenting queries with complementary descriptions~\cite{liu2023filling, momentGPT}.
Despite these prior efforts, ZMR still suffers from both gaps, often leading to unreliable span generation and scoring.
Therefore, we propose a novel framework that leverages video-intrinsic self-relationships to mitigate them jointly. \\

\noindent\textcolor{black}{\textbf{Self-similarity for temporal boundary modeling.} 
Self-similarity has been studied as a fundamental cue for capturing internal relational patterns in images and videos~\cite{shechtman2007matching}.
Building on this idea, recent works exploit temporal similarity to obtain pseudo temporal proposal during training.
PSVL~\cite{nam2021zero}, a training-based MR method, constructs a frame-wise self-similarity matrix to discover pseudo event regions and pairs them with pseudo queries, which are then used to train a localization model.
Similarly, another training-based video grounding framework~\cite{kim2023language} segments videos using a temporal similarity matrix and trains a grounding model by aligning the resulting pseudo proposals with pseudo language features in a pretrained vision-language space.
In contrast, we alleviate modality and language-style gaps in a training-free manner: we derive query-agnostic candidate spans from intra-video self-similarity and perform span scoring, thereby reducing reliance on unstable query-dependent similarity scores.
}


\section{Method}

\begin{figure*}[t]
    \centering
    \begin{center}
        \includegraphics[width=1.0\textwidth]{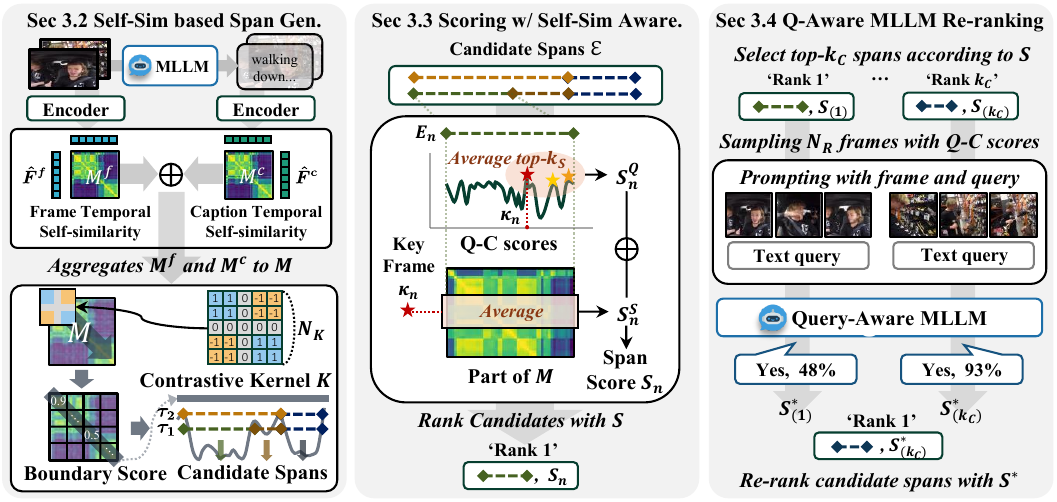}
    \end{center}    
    \vspace{-0.6cm}
    \caption{
    The overview of our Self-Similarity based Moment proposal and Scoring~(Self-SiMS). 
    First, features from the frames and captions are extracted to construct two distinct self-similarity maps, $M^f$ and $M^c$, which are then aggregated into a unified map, $M$.
    Each frame is then evaluated with a boundary score computed using a contrastive kernel, and candidate spans are generated in temporal order~(Sec. 3.2).
    Next, the spans are ranked based on query-matching span score $S^Q$, which is computed by averaging the top-$k_S$ query-caption scores.
    In addition, the self-matching span score $S^S$ is derived from the unified Self-similarity map $M$ and then combined with the former score~(Sec. 3.3).
    Finally, the MLLM-based reranking is applied to the top-ranked spans by asking whether each query-frame pair is relevant, and the probability of the 'Yes' answer is used as the score~(Sec. 3.4).
    }
    \label{fig:main}
    \vspace{-0.6cm}
\end{figure*}

\subsection{Overview}

Given an untrimmed video $V = {{\{v_i\}}_{i=1}^{L}}$ with $L$ sampled frames and a textual query $Q$, zero-shot video moment retrieval aims to localize the temporal span aligned with $Q$ without requiring any training. Fig.~\ref{fig:main} illustrates the overall pipeline of Self-SiMS. We first extract frame features $F^f$ and frame-wise caption features $F^c$, and use them to build two temporal self-similarity maps, $M^f$ and $M^c$. These maps are merged into a unified self-similarity map $M$, from which we compute boundary scores with a contrastive kernel and generate candidate spans. 
Each candidate is then scored by combining the Query-Matching Span Score $S^Q$, computed from the top-$k_S$ query--caption similarities within the span, and the Self-Matching Span Score $S^S$, which measures internal span consistency using $M$. The combined score $S$ ranks the candidates, and the top-$k_C$ spans are further re-ranked by an MLLM through a query-aware Yes/No verification. \\

\noindent{\textbf{Span Boundary Scores.}}
Inspired by the event boundary detection method proposed in ~\cite{uboco}, we calculate self-similarity to construct the Temporal Self-similarity Matrix~(TSM), $M\in\mathbb{R}^{L \times L}$.
First, we apply $\ell_2$-normalization to the extracted features, $F^f$ for frames and $F^c$ for captions, resulting in $\hat{F}^f$ and $\hat{F}^c$, respectively.
Then, the TSMs for the frames and captions are computed as $M^f=\hat{F}^f(\hat{F}^f)^\top$ and $M^c=\hat{F}^c(\hat{F}^c)^\top$.
To integrate both modalities, we construct the unified TSM $M$ by equally weighting $M^f$ and $M^c$, i.e., $M=\tfrac{1}{2}(M^f+M^c)$.

To evaluate whether a frame is likely to be a boundary, we examine a local patch of the self-similarity map around that frame. Intuitively, a good boundary separates two temporally coherent regions. Frames before the boundary should be similar to each other, frames after the boundary should also be similar to each other, but frames across the boundary should be less similar. The contrastive kernel captures this pattern by rewarding within-side similarities and penalizing cross-side similarities. Specifically, the kernel $K\in\mathbb{R}^{N_K \times N_K}$ has positive weights in the top-left and bottom-right quadrants, negative weights in the top-right and bottom-left quadrants, and zeros in the central row and column. The boundary score of the $i$-th frame is computed as
\begin{equation}
    b_i = P_i \odot K,
\end{equation}
where $P_i$ is an $N_K\times N_K$ patch of $M$ centered at the diagonal position $(i,i)$, and $\odot$ denotes element-wise multiplication. The TSM is zero-padded so that this operation can be applied uniformly to every frame. \\

\noindent{\textbf{Candidate Spans.}}  
To generate candidate spans from the boundary scores, we define a set of instance-wise dynamic thresholds~$\mathcal{T} = \{\tau_1, \tau_2, \dots, \tau_{N_\mathcal{T}}\}$, which adapt to the intra-video context variability measured from the frame features $F^f$. 
Based on this variability, videos with more diverse contexts are assigned lower thresholds for finer temporal partitioning, whereas more homogeneous videos receive higher thresholds for coarser segmentation.
This variability-aware design removes the need for manual threshold tuning and generalizes well across datasets.
Detailed formulations are provided in the Appendix.
For each $\tau_j \in \mathcal{T}$, we construct boundary set~$\mathcal{B}_j$ containing the indices of frames that are both above the threshold and correspond to local maxima of the boundary scores:
\begin{equation}
\mathcal{B}_j = \{\, i \;\mid\; b_i \geq \tau_j,\; b_i \geq b_{i-1},\; b_i \geq b_{i+1} \,\} \;\cup\; \{1, L\}.
\end{equation} 
Here, the two extreme frames, $1$ and $L$, are included to ensure that the candidate spans cover the entire duration of the video.
By the boundary set, we obtain a candidate span~$E_n$ formally defined as:  
\begin{equation}
E_n = \{i \;\mid\; e_n^s \le i \le e_n^e \}, \quad 1 \leq e_n^s < e_n^e \leq L,
\end{equation} 
where each pair $(e_n^s, e_n^e)$ is formed by two successive frame indices in the temporal order within the same boundary set $\mathcal{B}_j$, thereby ensuring that the resulting different spans do not overlap.
Finally, the set of all candidate spans across thresholds is formally denoted by
\begin{equation}
\mathcal{E} = \{E_1, E_2, \dots, E_{N_{s}}\},
\end{equation} 
where $N_{s}$ denotes the total number of candidate spans.
Note that each $E_n$ in this set refers to the corresponding candidate span and can be referred to as the “$n$-th span” in subsequent descriptions. \\

\subsection{Span Scoring with Self-Similarity Awareness}
\label{sec:Sapn_scoring}

To identify the relevant candidate span, each span should have its relevance score with the query. 
Previous approach~\cite{momentGPT} computes this score by averaging the frame-level similarities between the query and captions. 
However, such mean-based scoring is unreliable, as query–caption similarity often exhibits noisy and inconsistent distributions due to the language-style gap.

To mitigate this limitation, we use the top-$k_S$ frame-level similarities within the span for scoring, which are robust to the corruption induced by the language-style gap.
In addition, to enhance the reliability of scoring, we propose the Self-Matching Span Score, derived from TSM.
Finally, each candidate span is scored by combining the Query-Matching Span Score and the Self-Matching Span Score. \\

\noindent{\textbf{Query–Matching Span Score.}}
Given a textual query $Q$, we extract the query feature $F^q$ using the same encoder employed for the caption features $F^c$.
Then, the cosine similarity between the query feature and caption features is calculated to produce the frame-level scores $S^{f} \in \mathbb{R}^L$  as follows:
\begin{equation}
\label{eq:frame_score}
S^f_i = \frac{F^q \cdot F^c_i}{\|F^q\| \; \|F^c_i\|},
\end{equation}
where $F^c_i$ and $S^f_i$ denote the caption feature and frame-level score of the $i$-th frame, respectively.
The Query-Matching Span Score~(QMS), denoted as $S^Q$, is defined as the mean of the top-$k_S$ frame scores within each span.
Specifically, for the $n$-th span $E_n$, containing the indices from the start to the end frames $e^\text{s}_n$ and $e^\text{e}_n$, the score is computed as follows:
\begin{equation}
S^Q_n = \frac{1}{k_S} \sum_{i=e^\text{s}_n}^{e^\text{e}_n} \mathbf{1}\left[i \in I_{k_S}^{E_n}\right] S^f_i, 
\end{equation}
where $I_{k_S}^{E_n}$ is a set of indices with the top-$k_S$ highest frame-level scores defined in Eq.~(\ref{eq:frame_score}) within $E_n$.
This approach prioritizes the most relevant frames over the average, helping to mitigate noise in the query–caption similarity by using the most reliable scores. As a result, the span score more reliably reflects each span’s relevance to the query. \\

\noindent{\textbf{Self-Matching Span Score.}}
However, relying solely on a subset of sampled frames inherently limits the ability to capture alignment between the query and the full temporal context of a span.
To address this issue, we leverage the previously computed TSM used in span generation to evaluate contextual consistency within the span while reducing the influence of the language-style gap.
The Self-Matching Span Score (SMS), denoted as $S^{S}$, is defined as the mean similarity between the key frame, which is the frame with the highest frame-level score in the span, and all other frames within that span.
Let $\kappa_n$ denote the index of the key frame in the span $E_n$, that is,
\begin{equation}
\kappa_n = \arg\max_{i \in E_n} S^f_i.
\end{equation}
Then, SMS for the $n$-th candidate span is calculated as
\begin{equation}
S^S_n = \frac{1}{|E_n|} \sum_{j \in E_n} M_{\kappa_n, j},
\end{equation}
where $M_{i, j}$ denotes the element of the TSM $M$ at the position $(i, j)$.
As a result, candidate spans that are longer yet include inconsistent contexts are suppressed through reduced span scores. Finally, the total score $S_n$ of the candidate span is weighted sum of $S_n^Q$ and $S_n^S$ as follows:
\begin{equation}
S_n = (1-\alpha)\, S^Q_n + \alpha\, S^S_n, \quad \alpha \in [0,1],
\end{equation}
where $\alpha$ is a hyperparameter for the span score weights.

\subsection{Query-Aware Gap Reduction via MLLM Re-Ranking}
Recent remarkable advances in multimodal large language models (MLLMs) have demonstrated exceptionally strong cross-modal reasoning ability, making them effective for re-ranking tasks~\cite{lin2024mm_embed,chen2024mllm_reranker,sun2023rerank_1,jin2025rerank_2}.
Building on this, we further reduce the language-style gap between queries and MLLM-generated captions by refining candidate spans through MLLM-based re-ranking. \\

Concretely, we re-rank the top-$k_C$ candidate spans from the initial scoring stage, ranked according to their span score~$S$.
Let $S_{(m)}$ and $E_{(m)}$ denote the $m$-th highest span score and its corresponding candidate span, where $1 \le m \le k_C$.
Within each selected span, we sample $N_{R}$ representative frames according to their query–caption similarity score. 
Further details of the $N_R$ frame sampling process are provided in the Appendix.
Then, each representative frame is evaluated by the MLLM using a \textit{Yes/No} prompt, and the resulting logits are converted into probabilities via softmax. The resulting \textit{Yes} probability is used as $S^{r}_{i}$, denoting the frame-level re-ranking score for the $i$-th frame.
By averaging the top-$k_R$ highest values from the set of frame-level re-ranking scores $S_i^r$, we compute the span-level re-ranking score $S_{(m)}^R$.
Finally, $S_{(m)}^*$, the final score of $E_{(m)}$, is defined as a weighted combination of the initial span score $S_{(m)}$ and the MLLM-based re-ranking signal $S_{(m)}^R$:
\begin{equation}
S_{(m)}^* = (1-\beta)\, S_{(m)} + \beta\, S_{(m)}^R, \quad \beta \in [0,1],
\end{equation}
where $\beta$ is a weighting parameter.
This re-ranking process leverages the MLLM’s cross-modal reasoning to directly verify semantic alignment between the query and candidate captions, rather than relying solely on similarity scores, enabling query-aware reasoning and more reliable selection of the most relevant spans. 


\section{Experiments}


\subsection{Experimental Setup}

\noindent{\textbf{Datasets.}} 
We evaluate our method on four widely used benchmarks. QVHighlights~\cite{lei2021detecting} includes user-generated videos with fine-grained highlight annotations.
Charades-STA~\cite{gao2017tall} contains indoor daily activities paired with temporal sentence descriptions.
ActivityNet-Captions~\cite{krishna2017dense} covers diverse human activities with temporally aligned captions.
Additionally, we incorporate TVR~\cite{tvr}, which features long-form videos and complex natural language queries, to further validate the generalizability of our approach across varied video–language domains. \\

\noindent{\textbf{Metrics.}} 
Following prior zero-shot video moment retrieval~(ZMR) works, we evaluate our method using standard metrics tailored to each dataset. 
For QVHighlights, we report Recall@1 at IoU thresholds of 0.5 and 0.7, along with mAP@0.5 and the average mAP across multiple thresholds. 
These metrics jointly reflect the retrieval accuracy and the quality of segment ranking. 
For Charades-STA, ActivityNet-Captions, and TVR, we adopt R1@0.3/0.5/0.7 and mean IoU (mIoU), which evaluate both temporal localization precision and the alignment between predicted and ground-truth segments. \\

\noindent{\textbf{Implementation Details.}} We set the frame rates of QVHighlights, Charades-STA, ActivityNet-Captions, and TVR to 0.5, 1, 1, and 1 fps, respectively. Following prior work~\cite{momentGPT}, we use 0.5 fps for QVHighlights, and we set the remaining datasets to 1 fps for consistency.
We employ LLaMA-3.2-11B-Vision-Instruct~\cite{grattafiori2024llama3} for both video frame captioning and the final re-ranking stage. 
For scoring, the number of top scores is fixed at 3 for both initial span scoring~($k_S$) and re-ranking~($k_R$). 
We re-rank the top-$k_{C}$ candidate spans with $k_{C}=5$. 
In the re-ranking stage, $N_{R}$ frames are sampled from each span, set to half of the span length with lower and upper bounds of 10 and 30, respectively.
The weighting parameters for the final span score, $\alpha$ and $\beta$, are set to 0.1 and 0.5, respectively. 
All experiments are conducted on an NVIDIA RTX A6000 GPU. \\

\begin{table*}[t]
\centering
\begin{adjustbox}{width=\textwidth}
\begin{tabular}{lc|ccc |c|ccc|c}
\hhline{----------}
\multirow{2}{*}{Method} & \multirow{2}{*}{Setting} & \multicolumn{4}{c|}{QVHighlights test} & \multicolumn{4}{c}{QVHighlights val} \\
\hhline{~~|---|-|---|-}
 &  & R1@0.5 & R1@0.7 & mAP@0.5 & \multicolumn{1}{>{\columncolor{MyLightGray}}c|}{mAP@avg} & R1@0.5 & R1@0.7 & mAP@0.5 & \multicolumn{1}{>{\columncolor{MyLightGray}}c}{mAP@avg} \\
\hhline{----------}
Moment-DETR~\cite{lei2021detecting} & FS & 52.9 & 33.0 & 54.8 & {\cellcolor{MyLightGray}}30.7 & 54.2 & 33.4 & 55.4 & {\cellcolor{MyLightGray}}31.1 \\
UMT~\cite{liu2022umt} & FS & 56.4 & 40.8 & 53.1 & {\cellcolor{MyLightGray}}35.4 & - & - & - & {\cellcolor{MyLightGray}}- \\
\hhline{----------}
CPL~\cite{zheng2022bweakly} & WS & 30.8 & 10.8 & 22.8 & {\cellcolor{MyLightGray}}- & - & - & - & {\cellcolor{MyLightGray}}- \\
CPI~\cite{kong2023dynamic} & WS & 32.3 & 11.8 & 23.7 & {\cellcolor{MyLightGray}}- & - & - & - & {\cellcolor{MyLightGray}}- \\
\hhline{----------}
TFVTG\textsuperscript{\ddag}~\cite{TFVTG} & ZS & 21.6 & 8.0 & 20.1 & {\cellcolor{MyLightGray}}7.9 & 21.0 & 7.4 & 19.2 & {\cellcolor{MyLightGray}}7.3 \\

Moment-GPT~\cite{momentGPT} & ZS & {58.3} & {37.7} & {55.1} & {\cellcolor{MyLightGray}}{35.0} & {58.9} & {38.6} & {55.7} & {\cellcolor{MyLightGray}}{35.9} \\


\textbf{Self-SiMS~(Ours)} & ZS & \textbf{59.7} & \textbf{42.2} & \textbf{59.2} & {\cellcolor{MyLightGray}}\textbf{38.3} & \textbf{61.0} & \textbf{43.2} & \textbf{60.5} & {\cellcolor{MyLightGray}}\textbf{39.3} \\
\hhline{----------}
\end{tabular}
\end{adjustbox}
\vspace{0.1cm}
\caption{Comparison on QVHighlights test and validation sets. FS: Fully Supervised, WS: Weakly Supervised, ZS: Zero-Shot.  \ddag: Reproduced with the same LLM and FPS with ours.}
\label{tab:QVHighlights}
\vspace{-0.8cm}
\end{table*}

\begin{table*}[t]
\centering
\begin{adjustbox}{width=\textwidth}
\begin{tabular}{lc|ccc |c|ccc|c}
\hhline{----------}
\multirow{2}{*}{Method} & \multirow{2}{*}{Setting} & \multicolumn{4}{c|}{Charades-STA} & \multicolumn{4}{c}{ActivityNet-Captions} \\
\hhline{~~|----|----}
 &  & R1@0.3 & R1@0.5 & R1@0.7 & \multicolumn{1}{>{\columncolor{MyLightGray}}c|}{mIoU} & R1@0.3 & R1@0.5 & R1@0.7 & \multicolumn{1}{>{\columncolor{MyLightGray}}c}{mIoU} \\
\hhline{----------}
VTimeLLM~\cite{huang2024vtimellm} & FS & 55.3 & 34.3 & 14.7 & {\cellcolor{MyLightGray}}34.6 & 44.8 & 29.5 & 14.2 & {\cellcolor{MyLightGray}}31.4 \\
Moment-DETR~\cite{lei2021detecting} & FS & 62.1 & 48.2 & 25.3 & {\cellcolor{MyLightGray}}42.3 & 52.6 & 32.5 & 15.3 & {\cellcolor{MyLightGray}}37.8 \\
\hhline{----------}
CNM~\cite{zheng2022weakly} & WS & 50.0 & 36.2 & 14.2 & {\cellcolor{MyLightGray}}34.2 & 51.3 & 30.3 & 11.4 & {\cellcolor{MyLightGray}}33.9 \\
CPL~\cite{zheng2022bweakly} & WS & 56.0 & 38.1 & 20.3 & {\cellcolor{MyLightGray}}37.8 & 52.4 & 30.9 & 12.0 & {\cellcolor{MyLightGray}}32.6 \\
\hhline{----------}
TFVTG\textsuperscript{\dag}~\cite{TFVTG} & ZS & \underline{64.7} & 29.8 & 9.8 & {\cellcolor{MyLightGray}}38.2 & {49.9} & 26.8 & 12.6 & {\cellcolor{MyLightGray}}\underline{34.2} \\

TFVTG\textsuperscript{\ddag}~\cite{TFVTG} & ZS & \textbf{64.8} & 30.0 & 10.1 & {\cellcolor{MyLightGray}}\underline{38.4} & 49.5 & 26.5 & 12.3 & {\cellcolor{MyLightGray}}34.0 \\

Moment-GPT~\cite{momentGPT} & ZS & 58.2 & \underline{38.4} & \textbf{21.6} & {\cellcolor{MyLightGray}}36.5 & 48.1 & \textbf{31.1} & \textbf{14.9} & {\cellcolor{MyLightGray}}30.8 \\


\textbf{Self-SiMS~(Ours)} & ZS & {62.7} & \textbf{39.7} & \underline{21.0} & {\cellcolor{MyLightGray}}\textbf{41.9} & \textbf{49.9} & \underline{28.2} & \underline{13.8} & {\cellcolor{MyLightGray}}\textbf{34.7} \\

\hhline{----------}
\end{tabular}
\end{adjustbox}
\vspace{0.1cm}
\caption{Comparative evaluation on Charades-STA and ActivityNet-Captions. ZS: Zero-Shot. \dag: Reproduced results with the same FPS with ours. \ddag: Reproduced with the same LLM and FPS with ours.}
\label{tab:charades_activitynet}
\vspace{-0.4cm}
\end{table*}

\vspace{-0.2cm}
\begin{table}[t]
\centering
\begin{adjustbox}{}
\begin{tabular}{l | c c |c c c| c}
\hline
\multirow{2}{*}{Method}& \multicolumn{6}{c}{TVR}\\ \hhline{~|------}
& \textit{s} & \textit{f} & R1@0.3 & R1@0.5 & R1@0.7 & {\cellcolor{MyLightGray}}mIoU \\
\hline
 & 40 & 1.0 & 13.1 & 6.0 & 3.0 & {\cellcolor{MyLightGray}}15.1 \\
TFVTG\textsuperscript{\ddag}~\cite{TFVTG} & 50 & 0.5 & 14.4 & 6.9 & 2.7 & {\cellcolor{MyLightGray}}15.4 \\
 & 20 & 0.5 & 21.7 & 9.6 & 4.0 & {\cellcolor{MyLightGray}}18.6 \\
\hline
\textbf{Ours} & - & - &  \textbf{26.6} &\textbf{13.5} & \textbf{6.1} & {\cellcolor{MyLightGray}}\textbf{20.2} \\
\hline
\end{tabular}
\end{adjustbox}
\vspace{0.2cm}
\caption{Comparison on TVR validation dataset. For the baseline methods, \textit{s} (stride) and \textit{f} (max stride factor) are dataset-specific hyperparameters as reported in TFVTG~\cite{TFVTG}.}
\label{tab:tvr}
\vspace{-0.8cm}
\end{table}

\subsection{Comparison With the State-of-the-Arts}
Tabs.~\ref{tab:QVHighlights},\ref{tab:charades_activitynet}, and \ref{tab:tvr} report the performance of various ZMR methods across benchmark datasets.
In Tab.~\ref{tab:QVHighlights}, Self-SiMS outperforms existing ZMR baselines across all evaluation metrics on the QVHighlights. 
Note that TFVTG\textsuperscript{\ddag} was reproduced using the same large language model (LLM) employed in our framework, 
since TFVTG does not officially provide LLM rephrasing output for QVHighlights. 

Tab.~\ref{tab:charades_activitynet} presents the performance comparison on Charades-STA and ActivityNet-Captions. 
To ensure fairness, we reproduced TFVTG\textsuperscript{\dag} at 1 fps, 
as the original results were reported at 3 fps. 
TFVTG achieves relatively high R@0.3 and mIoU but exhibits clear drops at higher IoU thresholds (R1@0.5 and R1@0.7),
as its query–visual similarity often results in overly long spans.
Conversely, Moment-GPT performs better at higher thresholds but worse at R@0.3 and mIoU,
indicating that its query–caption similarity produces precise yet unstable short matches.
In contrast, our Self-SiMS overcomes these issues by producing more consistent spans aligned with the ground-truth, 
achieving the highest mIoU and more reliable temporal localization while maintaining competitive recall. 

Tab.~\ref{tab:tvr} summarizes the evaluation on the TVR dataset. 
As the official implementation of Moment-GPT is unavailable, we omit its results for fairness. 
For TFVTG, we report multiple configurations using the hyperparameters $s$ (stride) and $f$ (max stride factor) 
as specified in the original paper, ensuring a fair comparison. 
Our Self-SiMS consistently surpasses the baseline across all parameter settings, marking a substantial improvement. Moreover, while the baseline requires dataset-specific hyperparameter tuning, Self-SiMS delivers strong performance with a single configuration, highlighting its robustness.

\subsection{Ablation Studies}

\noindent{\textbf{Length of Candidate Spans Analysis.}}

\begin{figure*}[t]
    \centering
    \begin{center}
    \includegraphics[width=0.7\textwidth]{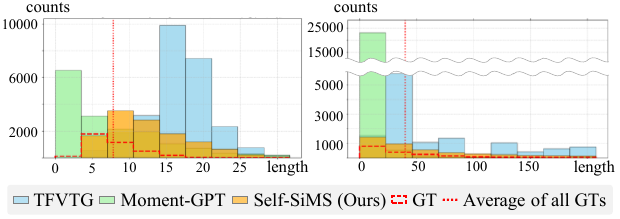}
            \end{center}    
            \vspace{-0.6cm}
    \caption{Histogram of all span lengths on Charades-STA (left) and ActivityNet-Captions (right).}
    \label{fig:span_len}
    \vspace{-0.6cm}
\end{figure*}

Fig.~\ref{fig:span_len} illustrates the candidate span length distributions on Charades-STA (left) and ActivityNet-Captions (right).
The ground-truth (GT) distribution is indicated by the red dashed line.
This analysis helps explain the performance differences observed on these two datasets, particularly the recall tendencies in Tab.~\ref{tab:charades_activitynet}.
TFVTG, affected by the modality gap, produces uniform similarity scores with weak temporal distinction. Consequently, it often produces excessively long spans, achieving relatively high R@0.3 but exhibiting clear degradation under stricter IoU thresholds.
Conversely, Moment-GPT suffers from a language-style gap, leading to noisy and unstable similarity score distribution. These sharp, unstable peaks often yield overly short spans that perform better under tight criteria such as R@0.7 yet fail to capture full GT intervals.
In contrast, our Self-SiMS generates balanced spans that align more closely with the GT distribution by leveraging self-similarity free from such modality and style gaps, resulting in higher mIoU and more consistent temporal localization performance. \\

\begin{table}[t]
    \centering
    \begin{minipage}[t]{0.45\linewidth}
        \centering\vspace{10pt}
        \begin{adjustbox}{width=\linewidth}
\begin{tabular}{l c c}
\hline
Method & Oracle-mIoU (\%) & Avg. \# Span Cand.  \\
\hline
TFVTG & 38.16 & 8.38  \\
Moment-GPT & 65.01 & 7.62  \\
\hline
Self-SiMS (Ours) & \textbf{71.13} & \textbf{6.46} \\
\hline
\end{tabular}
        \end{adjustbox}
        \vspace{0.1cm}
        \caption{Comparison of Oracle mIoU on the QVHighlights validation set. Avg. \# Span Cand. denotes the average number of span candidates across on dataset.}
        \label{tab:ablation_study1_1}
    \end{minipage}\hfill
    \begin{minipage}[t]{0.52\linewidth}
        \centering\vspace{0pt}
        \includegraphics[width=\linewidth]{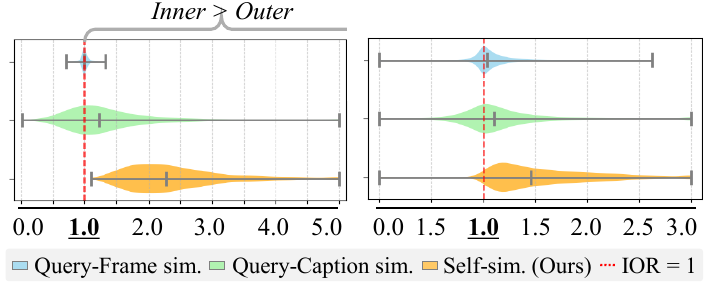}
        
        \captionof{figure}{Inner-to-Outer Ratio (IOR) distributions on Charades (left) and ActivityNet-Captions (right).}
        \label{fig:IOR}
    \end{minipage}
    \vspace{-0.8cm}
\end{table}

\noindent{\textbf{Effectiveness of Span Generator.}} 
To validate the effectiveness of our span generator, we evaluate Oracle mIoU on the QVHighlights, comparing against TFVTG~\cite{TFVTG} and Moment-GPT~\cite{momentGPT}. 
Oracle mIoU is defined as the mean IoU obtained by selecting the candidate span with the maximum overlap with the ground truth, reflecting the quality of the span candidates. 
As shown in Tab.~\ref{tab:ablation_study1_1}, our method achieves an Oracle mIoU of 71.13, outperforming both baselines. 
Moreover, although this performance generally increases with a larger number of candidate spans, our method achieves the best results despite requiring the lowest average number of candidates~(6.46).
This improvement demonstrates that our method generates more reliable spans by capturing the intrinsic structure of the video while avoiding modality and language-style gaps. \\

\noindent{\textbf{Inner-to-Outer Ratio Distribution.}} Fig.~{\ref{fig:IOR}} reports IOR distributions on Charades-STA and ActivityNet-Captions, where higher IOR values indicate stronger separation between the ground-truth span and surrounding background.
Self-SiMS shows higher and more stable IOR values, demonstrating that its self-similarity-based boundary scores capture temporal segments more reliably and lead to robust localization. \\

\noindent{\textbf{Scoring Component Analysis.}} 
We conduct an ablation study on the QVHighlights validation set to analyze the contribution of each component in our proposed scoring and re-ranking pipeline, with the results presented in Tab.~\ref{tab:ablation_study2}.
The baseline adopts naive mean scoring, which averages all frame-level scores within a span, as in prior methods. 
Replacing mean scoring with the Query-Matching Span Score~(QMS) yields improvements across all metrics, supporting our hypothesis that employing the most relevant signals is more effective than relying on the average of noisy query-caption scores.
Adding the Self-Matching Span Score~(SMS) on QMS further increases performance, particularly for recall-based metrics such as R1@0.5 and R1@0.7, indicating that incorporating the entire span context through self-similarity enables more precise localization of ground-truth moments. 
Finally, incorporating MLLM re-ranking on top of QMS with SMS yields additional gains across all metrics by leveraging the MLLM’s ability to understand cross-modal information. 
This demonstrates that each component of the pipeline contributes effectively to the overall performance. \\ \\

\vspace{-0.4cm}
\begin{table}[t]
    \centering
    \begin{minipage}[t]{0.45\linewidth}
        \begin{adjustbox}{width=\linewidth}
\begin{tabular}{l c c c c}
\hline
Method & R1@0.5 & R1@0.7 & mAP@0.5 & mAP@avg \\
\hline
Mean scoring & 55.7 & 40.9 & 56.7 & 37.1 \\
\hline
QMS & 59.5 & 42.1 & 59.4 & 38.6 \\
\  + SMS & 60.3 & 42.9 & 59.6 & 38.7 \\
\quad + Re-ranking & \textbf{61.0} & \textbf{43.2} & \textbf{60.5} & \textbf{39.3} \\
\hline
\end{tabular}
\vspace{0.1cm}
        \end{adjustbox}
        \caption{Scoring component analysis of our scoring methods. QMS denotes the Query-Matching Score, and SMS denotes the Self-Matching Span Score.}
        \label{tab:ablation_study2}
    \end{minipage}\hfill
    \begin{minipage}[t]{0.52\linewidth}
        \begin{adjustbox}{width=\linewidth}
\begin{tabular}{l l l c c}
\hline
\textbf{Model} & \textbf{Component} & \textbf{MLLM / LLM} & \textbf{\#Params} & \textbf{Runtime} \\
\hline
\multirow{1}{*}{TFVTG} 
& Rephrasing & GPT4-Turbo~\cite{gpt4} & $\geq 100$B (est.) & -- \\
\hline

\multirow{3}{*}{Moment-GPT}
& Rephrasing & LLaMA3~\cite{grattafiori2024llama3} & 8B & 6.3s \\
& Captioning & MiniGPT-v2~\cite{minigptv2} & 7B & 249.4s \\
& Reranking & V-ChatGPT~\cite{video_chatgpt} & 7B & 7.4s \\

\hline

\multirow{2}{*}{Ours}
& Captioning & LLaMA3.2-V~\cite{grattafiori2024llama3} & 11B & 140.7s \\
& Reranking & LLaMA3.2-V~\cite{grattafiori2024llama3} & 11B & 13.0s \\
\cline{2-5}
& Captioning & MiniGPT-v2~\cite{minigptv2} & 7B & 249.4s \\
& Reranking & MiniGPT-v2~\cite{minigptv2} & 7B & 9.7s 
\\
\hline

\end{tabular}
\vspace{-0.6cm}

        \end{adjustbox}
        \textcolor{blue}{\caption{Comparison of MLLM/LLM usage, model scale, and per-video runtime across TFVTG, Moment-GPT, and our method.}
        \label{tab:llm_runtime_comparison}}
    \end{minipage}
    \vspace{-0.8cm}
\end{table}

\noindent\textbf{Complexity comparison.}
We compare the per-video runtime of TFVTG~\cite{TFVTG}, Moment-GPT~\cite{momentGPT}, and our method using 100 randomly sampled videos from Charades-STA~\cite{gao2017tall}.
Since non-MLLM/LLM components, such as our scoring stage, are lightweight (0.194s per video), we focus on MLLM/LLM-dependent stages and use a frame batch size of 1 for both captioning and reranking.
Tab.~\ref{tab:llm_runtime_comparison} summarizes the model configurations and runtimes.
TFVTG uses GPT-4 Turbo~\cite{gpt4} for query rephrasing. Its runtime cannot be directly measured because it is accessed through a commercial API, and its model size is estimated to exceed 100B parameters.
Moment-GPT employs three separate models for query rephrasing, captioning, and reranking, resulting in a total footprint of 22B parameters.
In contrast, our default setting uses a single LLaMA-3.2-Vision~\cite{grattafiori2024llama3} model family for both captioning and reranking, reducing the model footprint to 11B parameters.
Furthermore, reranking is computationally efficient because it evaluates only a small number of sampled representative frames.
To ensure a fair comparison under the same MLLM, we additionally replace LLaMA-3.2-Vision with MiniGPT-v2, which is also adopted by Moment-GPT.
Under this setting, the captioning cost is identical to that of Moment-GPT, while the reranking stage is slightly faster than when using LLaMA-3.2-Vision.
Detailed performance results obtained with MiniGPT-v2 are provided in the supplementary MLLM ablation study.
Notably, our method achieves comparable or superior performance while using the same MLLM backbone.

\begin{figure*}[t]
    \centering
    \begin{center}
        \includegraphics[width=1.0\textwidth]{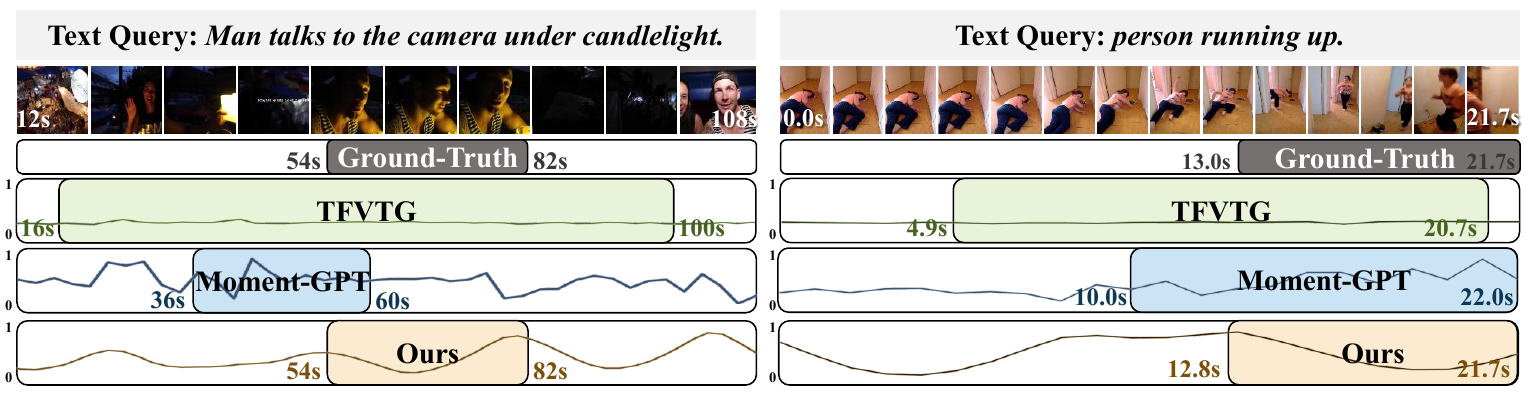}
    \end{center}    
    \vspace{-0.6cm}
    \caption{Qualitative results on QVHighlights (left) and Charades-STA (right).}
    \vspace{-0.6cm}
    \label{fig:qualitative}
\end{figure*}

\subsection{Qualitative results}
We provide a qualitative comparison in Fig.~\ref{fig:qualitative}, illustrating span localization results, TFVTG~\cite{TFVTG}, Moment-GPT~\cite{momentGPT}, and our method, along with the corresponding span candidate score distributions.
TFVTG, which relies on query–visual frame similarity, generates overly long and inaccurate spans due to the modality gap that hinders fine-grained semantic alignment.
Moment-GPT, which instead depends on query–caption similarity, suffers from a language-style gap that produces inconsistent similarity maps and hampers accurate localization.
In contrast, Self-SiMS leverages boundary scores derived from the self-similarity, enabling robust localization of the correct spans.

\section{Conclusion}

In this paper, we first define the modality and language-style gaps in zero-shot video moment retrieval (ZMR), and quantify their impact using the Inner-to-Outer Ratio (IOR), which measures how well a similarity score separates the ground-truth span from the surrounding background.
To mitigate these gaps, we propose Self-Similarity-based Moment Proposal and Scoring (Self-SiMS), which leverages intra-video self-similarity for both span generation and scoring.
Experiments and ablations show that Self-SiMS mitigates errors from unreliable query-dependent similarity and achieves robust performance, demonstrating self-similarity as an effective cue for training-free ZMR.



\section*{Acknowledgements}
This work was supported in part by MSIT/IITP (No. RS-2022-II220680, RS-2020-II201821, RS-2019-II190421, RS-2024-00459618, RS-2024-00360227, RS-2024-00437633, RS-2024-00437102, RS-2025-25442569), MSIT/NRF (No. RS-2024-00357729), KNPA/KIPoT (No. RS-2025-25393280), and SEMES-SKKU collaboration funded by SEMES.

\clearpage
%
%
\bibliographystyle{splncs04}
\bibliography{main}

\clearpage
\setcounter{page}{1}

{
\centering
\Large
\textbf{Mitigating Modality and Language-Style Gaps \\for Zero-Shot Video Moment Retrieval}\\
\vspace{0.5em}
Supplementary Material \\
}


\section{Additional Method Details}

\subsection{Definition of Threshold for Generating Candidate Spans}
In this section, we define the instance-wise dynamic threshold $\tau_j$, which suppresses the frames with low boundary scores.
The motivation for adopting a dynamic threshold is that segmenting videos into contextually consistent spans is highly sensitive to the variability of the context within the video.
A fixed threshold, which ignores this factor, fails to segment the videos appropriately.
Therefore, we propose the set of thresholds~$\mathcal{T} = \{\tau_1, \tau_2, \dots, \tau_{N_\mathcal{T}}\}$ which adapt to the video context variability. \\

\noindent\textbf{Video Context Variability.}
We introduce an indicator of intra-video context variability to make the threshold respond to it.
First, we $\ell_2$-normalize the frame features $F^f\in\mathbb{R}^{L\times D^f}$ into $\hat{F}^f$.
Then, we obtain the mean of the normalized frame features $\mu^f$ as follows:
\begin{equation}
    \mu^f=\frac{1}{L}\sum_{i=1}^{L}\hat{F}^f_i.
\end{equation}
Then, we apply zero-centering on $\hat{F}_i^f$, that is, $\Tilde{F}_i^f=\hat{F}_i^f-\mu^f$.
We define the variability of the video context $\mathcal{V}$ by averaging the $\ell_2$-norm of zero-centered $\Tilde{F}_i^f$:
\begin{equation}
    \mathcal{V}=\frac{1}{L}\sum_{i=1}^{L}\|\Tilde{F}_i^f\|_2.
\end{equation}
The above formulations can be interpreted as follows: if the original frame features $F^f$ are uniformly spread over an arbitrary hypersphere, the mean vector $\mu^f$ computed after $\ell_2$-normalization will be close to the zero vector, which in turn makes $\mathcal{V}$ close to 1.
Conversely, if the frame features are biased toward a particular direction, which means that some frame features have similar context, the mean vector $\mu^f$ will also be shifted toward that direction, resulting in a smaller norm of $\Tilde{F}^f$ and consequently a lower $\mathcal{V}$.
In other words, $\mathcal{V}$ reflects the diversity of the contexts contained in a video instance.
Therefore, we regard $\mathcal{V}$ as an indicator of the intra-video context variability.\\

\noindent\textbf{Definition of Threshold.}
Next, we define the threshold $\tau_j$ based on the video context variability.
It is obvious that videos with diverse contexts should undergo fine-grained segmentation
whereas homogeneous ones require coarse-grained segmentation.
Thus, the threshold should decrease as the indicator $\mathcal{V}$ grows, that is, leading to more candidate spans by increasing the size of boundary set $\mathcal{B}_j$.
In contrast, the threshold should grow as the indicator $\mathcal{V}$ decreases, thereby reducing the number of candidate spans by suppressing more frames.
Therefore, we specify the threshold as inversely related to the variability:
\begin{equation}
    \tau_j=1-\lambda_j\mathcal{V} \quad \text{where} ~\lambda_j\in\Lambda_{j=1}^{N_\tau},
\end{equation}
where $\Lambda$ is a set of constants that control threshold scaling and the generation of spans with diverse lengths.
In practice, we set $\Lambda$ to $\{1,2,3\}$ across the benchmark datasets, hence $N_\tau=3$.
Note that the threshold changes instance-wise via $\mathcal{V}$.

\subsection{Detailed Procedure of MLLM-Based Re-Ranking}
In this section, we provide a more detailed explanation of the MLLM-based re-ranking process, which complements the concise description in the main paper. \\

\noindent\textbf{Candidate Selection.}
For efficiency, re-ranking is applied only to the top-$k_{C}$ candidate spans from the initial scoring stage, as these are the most likely to contain the correct answer. In our implementation, we set $k_{C}=5$, balancing computational efficiency with sufficient coverage of promising spans. The selected candidate spans are represented as
\begin{equation}
\mathcal{E}_{R} = \{ [e_n^{\text{s}}, e_n^{\text{e}}] \}_{n=1}^{k_C},
\end{equation}
where $e_n^{\text{s}}$ and $e_n^{\text{e}}$ denote the start and end frame indices of the $n$-th selected span, forming the set of spans subject to re-ranking.
For the $n$-th span $[e_n^{\text{s}}, e_n^{\text{e}}]$, the number of frames is defined as
\begin{equation}
N^{f}_{n} = e_n^{\text{e}} - e_n^{\text{s}} + 1.
\end{equation} \\

\noindent\textbf{Frame Sampling.}
For each span, our goal is to extract representative frames that best capture the semantic content relevant to the query. To achieve this, we use the previously computed query-caption~(QC) scores $S^f$, which measure the alignment between the query and individual video frames. Frames with higher QC scores are prioritized for sampling, as they are assumed to be more informative for query verification. The number of sampled frames is determined adaptively according to the span length with a sampling rate $\rho$, and is clamped between a minimum and maximum to avoid undersampling short spans or oversampling long spans:
\begin{equation}
N_{R} = \max\!\big(N^{f}_{\min}, \min( \rho \cdot N^{f}_{n}, N^{f}_{\max})\big),
\end{equation}
where $N^f_{\min}$ and $N^f_{\max}$ are 10 and 30, respectively.
The $N_{R}$ frames with the highest QC scores are retained, and their indices are sorted in temporal order to preserve sequential structure. \\

\noindent\textbf{Prompting and Scoring.}
The sampled frames are passed to the MLLM for comparison with the text query. To ensure consistent behavior, we employ a structured prompt that asks the model whether the query is aligned with the input frame. The prompt is designed as:

\begin{quote}
\textit{``You are a visual fact-checker. Your job is to verify a statement against a single video frame.
Do not infer or guess information that is not clearly visible. Statement to verify:} \textbf{[Query]}.
\textit{Based strictly on the visual information in the image, is this statement true? Answer with only `Yes' or `No'.''}
\end{quote}

Here, \textbf{[Query]} denotes the placeholder where the text query is inserted.
The MLLM produces logits for both ``Yes'' and ``No.'' Let $l^{\text{yes}}_i$ and $l^{\text{no}}_i$ denote the logits corresponding to $i$-th frame. The frame-level re-ranking score is defined as the softmax probability of the ``Yes'' token:
\begin{equation}
S^r_i = \frac{\exp(l^{\text{yes}}_i)}{\exp(l^{\text{yes}}_i) + \exp(l^{\text{no}}_i)}.
\end{equation} \\

\noindent\textbf{Span-Level Score Aggregation.}
To obtain a re-ranking score for an entire span, we aggregate the frame-level scores. Instead of averaging all sampled frames, which may include noisy or less informative ones, we select the top-$k_R$ verification scores~(with $k_{R}=3$) and compute their mean:
\begin{equation}
S^R_n = \frac{1}{k_R} \sum_{i=e^\text{s}_n}^{e^\text{e}_n} \mathbf{1}\left[i \in I_{k_R}^{E_n}\right] S^r_i,
\end{equation}
where $I_{k_R}^{E_n}$ is a set of indices with the top-$k_R$ highest frame-level re-ranking scores within the $n$-th candidate span $E_n$.
This design emphasizes the most confident evidence while reducing the impact of outliers.

\section{Additional Analyses}

\subsection{Hyperparameter analysis}

\begin{figure}[t]
    \centering
    \includegraphics[width=0.9\linewidth]{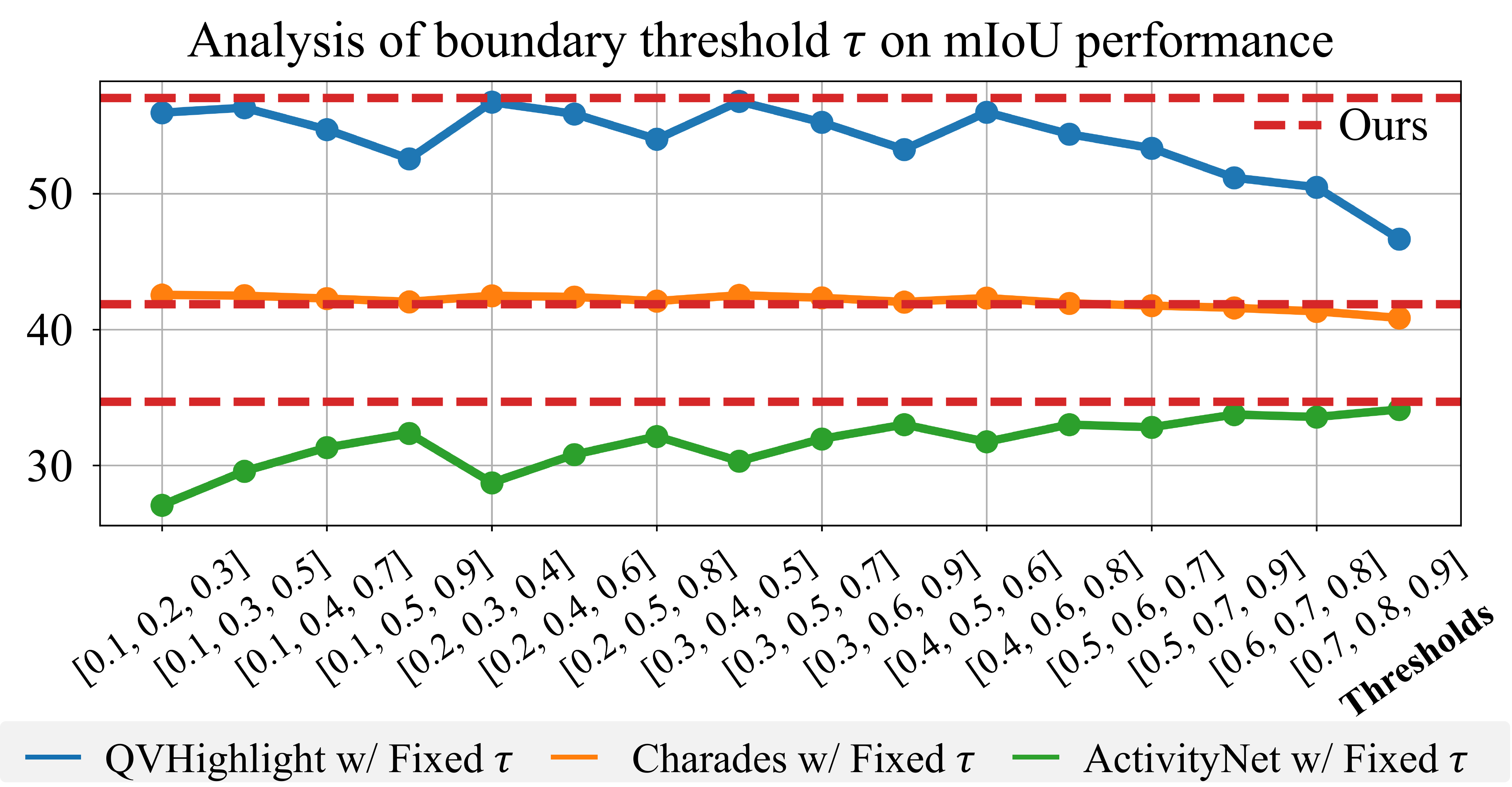}
    \caption{
mIoU performance comparison between fixed thresholds and our dynamic threshold across three datasets.
}

    \label{fig:tau_analysis}
\end{figure}

\begin{figure}[t]
    \centering
    \includegraphics[width=0.7\linewidth]{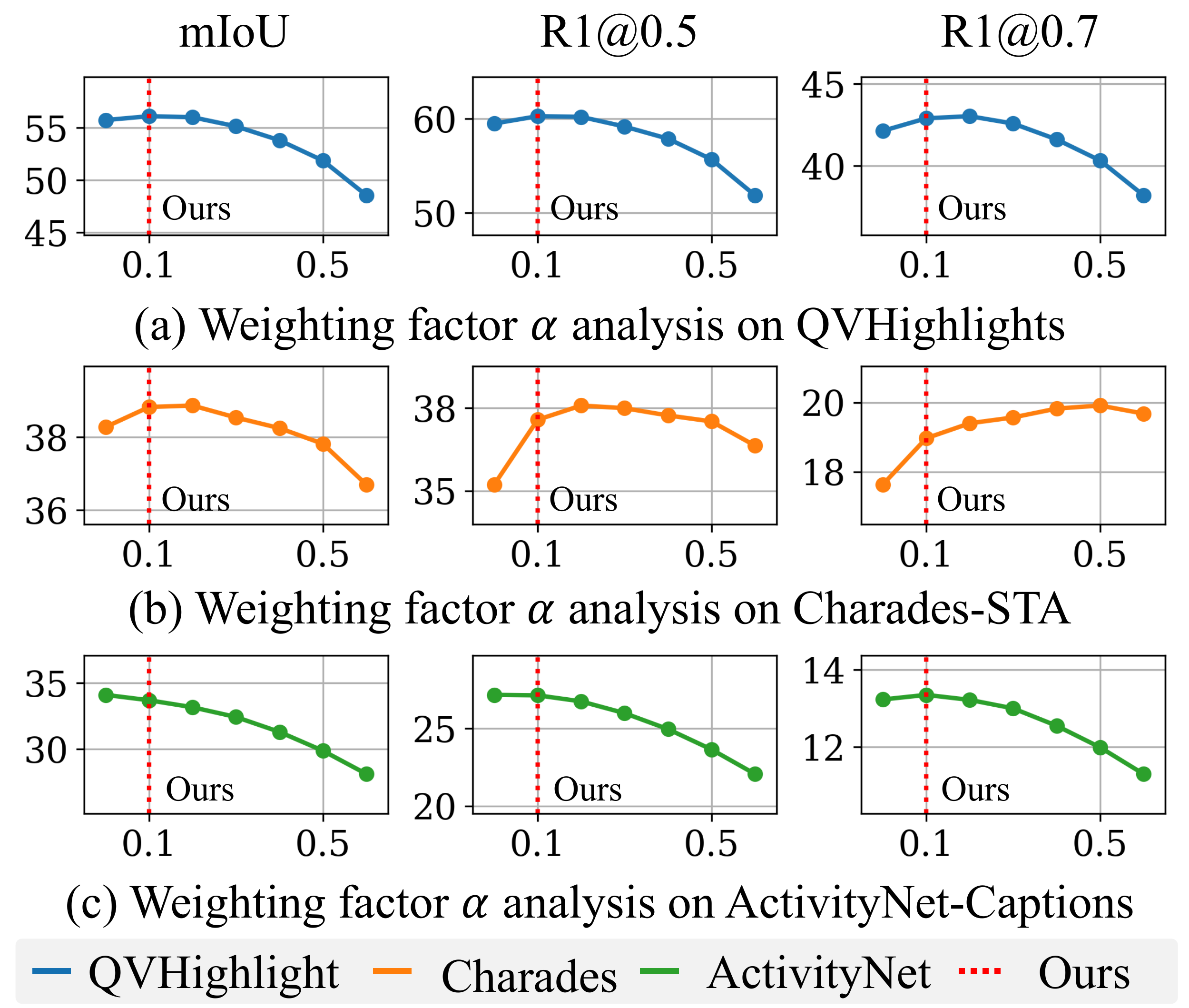}
    \caption{
Effect of weighting factor $\alpha$ on mIoU and recall across three datasets.
}
    \label{fig:alpha_analysis}
\end{figure}

In this section, we analyze the key hyperparameters of Self-SiMS, including \textcolor{black}{the boundary score thresholds} $\tau$, the top-$k_S$ selection strategy, and the weighting factors $\alpha$ and $\beta$. We examine how each component influences span generation and scoring. \\

\noindent\textbf{Boundary threshold $\mathcal{T}$.}
We analyze the effectiveness of the dynamic thresholds defined in $\mathcal{T}$ by comparing them with fixed threshold values.
Fig.~\ref{fig:tau_analysis} reports the performance of varying fixed thresholds~(solid lines) against the instance-wise dynamic threshold determined by ours~(red lines).
For fair comparison, each fixed-threshold setting is also evaluated as a set of three thresholds, matching the structure of our dynamic threshold set $\mathcal{T}$.
Across all datasets, the dynamic threshold consistently achieves similar or higher mIoU than any fixed threshold choice, demonstrating that adapting the threshold to the intra-video context leads to more reliable boundary detection.
This confirms that videos with diverse contexts benefit from finer segmentation, while more homogeneous videos require coarser segmentation, validating the motivation behind incorporating $\mathcal{V}$ into the boundary threshold design. \\

\noindent\textbf{Self-matching span score weighting factor $\alpha$.}
\textcolor{black}{For the initial span score, the weighting factor~$\alpha$ controls the contribution of the self-matching span score~(SMS) relative to the query-matching span score~(QMS).
Fig.~\ref{fig:alpha_analysis} illustrates the effect of varying $\alpha$.}
To fairly evaluate the contribution of SMS itself, this analysis excludes the MLLM re-ranking score and examines performance using only the combination of QMS and SMS.

Across the three datasets, we observe generally consistent trends.
Performance drops when $\alpha$ becomes too small, indicating that relying only on QMS without sufficient SMS contribution fails to capture the intra-span contextual consistency provided by self-similarity.
On the other hand, excessively large $\alpha$ also degrades performance because overly emphasizing SMS reduces the influence of query-dependent signals that are essential for accurate semantic alignment.

A moderate weighting around $\alpha = 0.1$ yields strong and stable results across all benchmarks, as indicated by the red dashed lines.
This shows that balanced integration of query-dependent and self-similarity-based signals results in robust span scoring, and that the model remains stable without high sensitivity to $\alpha$ within this range. \\

\begin{figure}[t]
    \centering
    \includegraphics[width=0.7\linewidth]{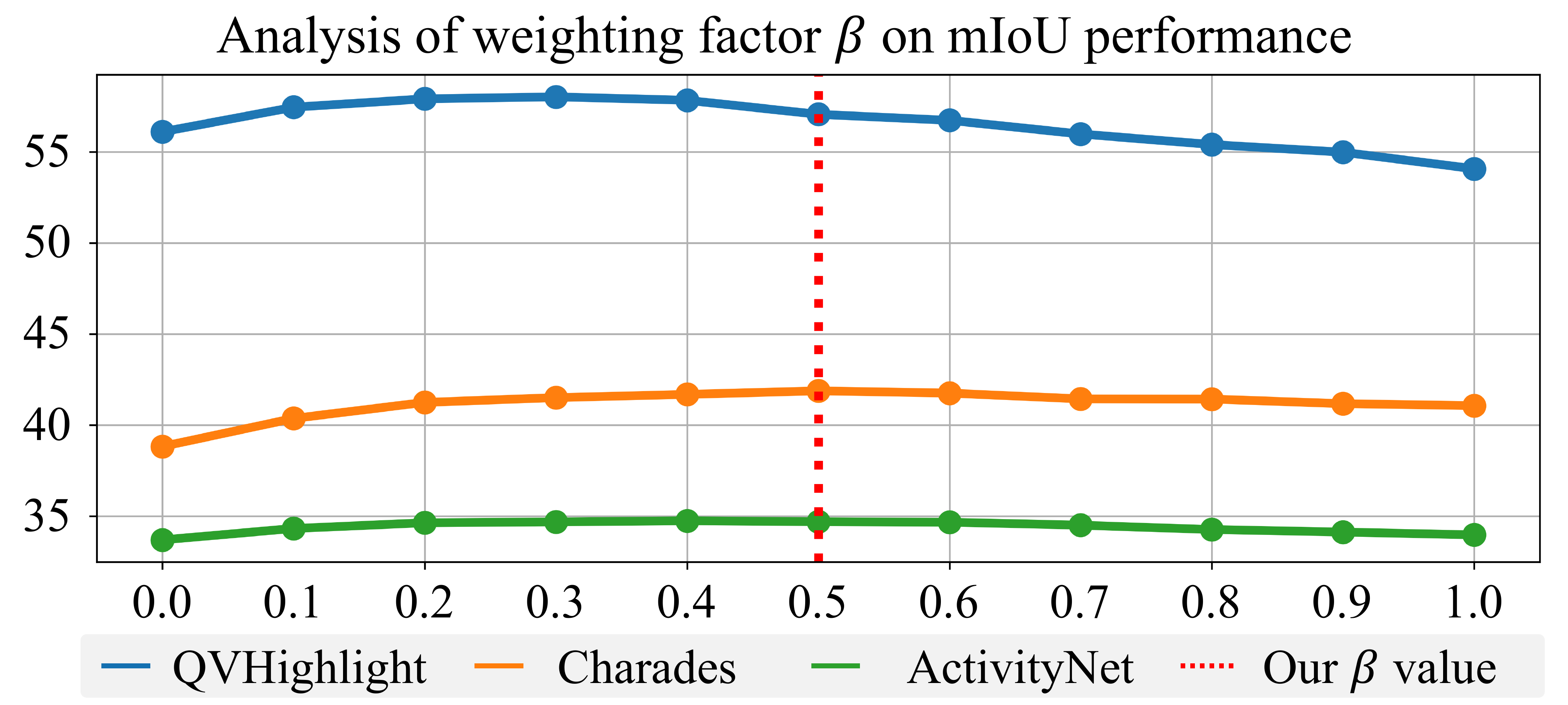}
    \caption{
Effect of weighting factor $\beta$ on mIoU for QVHighlights, Charades-STA, and ActivityNet-Captions.
}
    \label{fig:beta_analysis}
\end{figure}

\noindent\textbf{Reranking score weighting factor $\beta$.}
\textcolor{black}{For the final span score, the weighting factor~$\beta$ controls the contribution of the rereanked span score~$S_{(m)}^R$ relative to the initial span score~$S_{(m)}$.}
Fig.~\ref{fig:beta_analysis} illustrates how varying the weighting factor $\beta$ affects performance across different datasets.
We set the weighting factor $\beta$ to 0.5 so that the re-ranking mechanism generalizes consistently across datasets, where $\beta = 0.5$ corresponds to taking the average of the initial score $S$ and the MLLM re-ranking score.
As shown in the ablation results, $\beta = 0.5$ provides a stable choice, exhibiting high performance on all benchmarks with minimal sensitivity in its neighborhood.
Additionally, because the re-ranking stage evaluates only a small set of sampled frames rather than the full span due to computational cost, relying solely on the MLLM re-ranking score is not ideal.
For this reason, we combine the initial score $S$ with the MLLM re-ranking score, which provides a practical and effective balance between performance and efficiency. \\

\begin{figure}[t]
    \centering
    \includegraphics[width=0.7\linewidth]{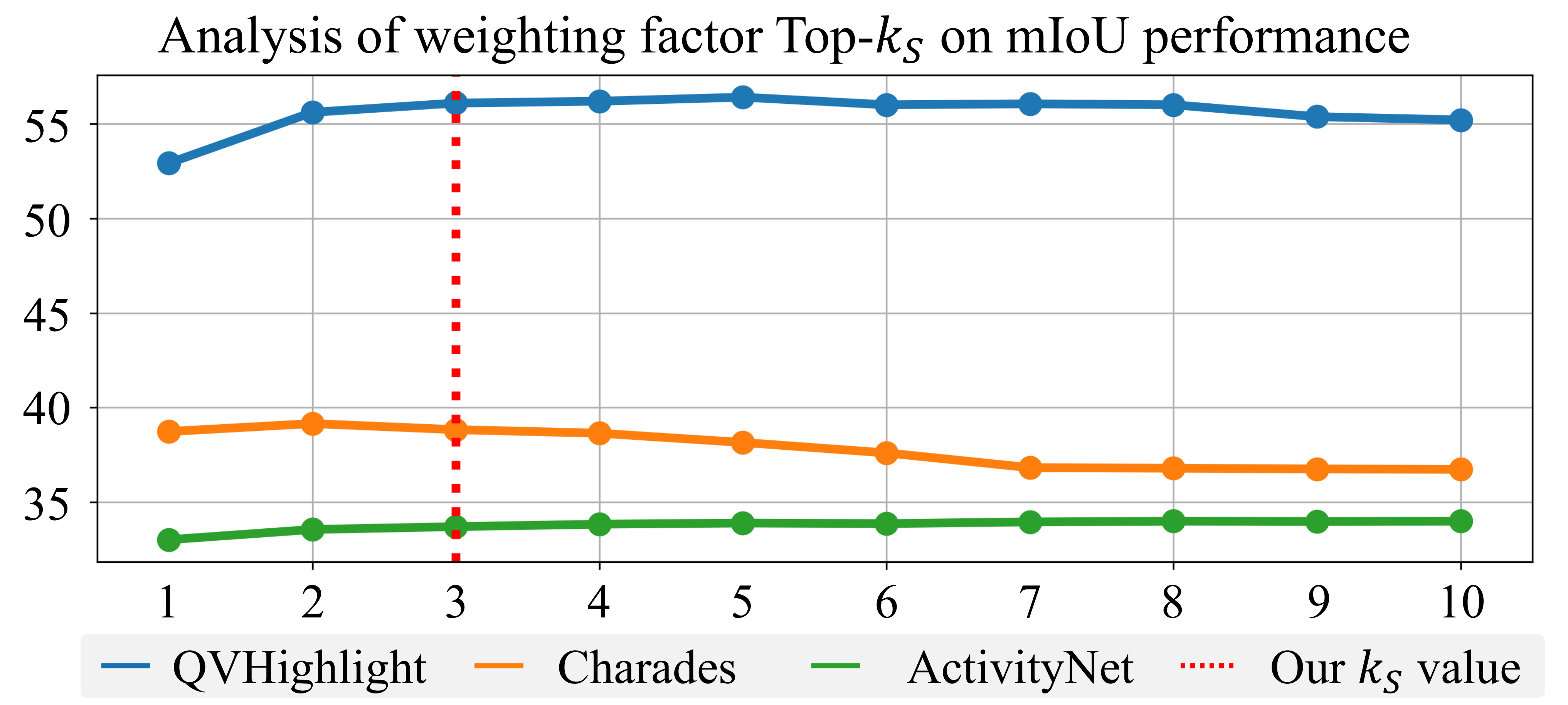}
    \caption{
Effect of Top-$k_S$ selection on mIoU for QVHighlights, Charades-STA, and ActivityNet-Captions.
}

    \label{fig:topk_s_analysis}
\end{figure}

\noindent\textbf{Analysis of the Top-$k_S$ Parameter.}
Figure~\ref{fig:topk_s_analysis} shows the effect of varying the top-$k_S$ parameter used in the Query-Matching Span Score.
When $k_S = 1$, the score is determined by only a single frame, resulting in insufficient contextual evidence within the span and consequently lower performance.
Conversely, as $k_S$ becomes too large, an increasing number of noisy query-caption scores are included, which dilutes the contribution of highly relevant frames and leads to performance degradation.
Across all datasets, we observe that values in the range of $k_S = 2$ to $4$ consistently achieve the best results.
Within this range, performance remains stable with minimal variation, indicating that the proposed scoring method operates robustly without being overly sensitive to the specific choice of $k_S$. \\


\begin{table*}[t]
\centering
\renewcommand{\arraystretch}{1.2}
\begin{adjustbox}{width=\textwidth}
\begin{tabular}{l | c c c c | c c c c | c c c c}
\hline
\multirow{2}{*}{\textbf{Method}} 
& \multicolumn{4}{c|}{\textbf{QVHighlights val}} 
& \multicolumn{4}{c|}{\textbf{Charades-STA}} 
& \multicolumn{4}{c}{\textbf{ActivityNet-Captions}} \\
\cline{2-13}
& {mAP} & {mAP@0.5} & {R1@0.5} & {R1@0.7}
& {mIoU} & {R1@0.3} & {R1@0.5} & {R1@0.7}
& {mIoU} & {R1@0.3} & {R1@0.5} & {R1@0.7} \\
\hline
\multicolumn{13}{l}{\textit{Before re-ranking}} \\
\hline
frame-only   
& 38.2 & 59.5 & 60.0 & 42.1
& 38.6 & 58.6 & 36.6 & 18.7
& 32.5 & 47.2 & 26.6 & 12.5 \\

caption-only 
& 35.6 & 57.0 & 57.0 & 39.0
& 38.6 & 58.6 & 36.2 & \textbf{18.7}
& 33.7 & \textbf{48.9} & \textbf{27.6} & \textbf{13.4} \\

\textbf{fused}        
& \textbf{38.7} & \textbf{59.6} & \textbf{60.3} & \textbf{42.9}
& \textbf{38.8} & \textbf{58.7} & \textbf{37.2} & \textbf{18.7}
& \textbf{33.7} & 48.5 & 27.1 & 13.4 \\
\hline
\multicolumn{13}{l}{\textit{After re-ranking}} \\
\hline
frame-only   
& 39.0 & \textbf{61.0} & \textbf{61.7} & \textbf{43.2}
& 41.7 & {62.9} & {39.7} & 20.8
& 33.7 & 49.1 & 27.7 & 12.9 \\

caption-only 
& 36.2 & 58.0 & 57.8 & 39.6
& 41.5 & \textbf{62.9} & 39.2 & 20.2
& \textbf{34.7} & 50.5 & \textbf{28.5} & \textbf{13.9} \\

\textbf{fused}        
& \textbf{39.3} & 60.5 & 61.0 & \textbf{43.2}
& \textbf{41.9} & 62.7 & \textbf{39.7} & \textbf{21.0}
& \textbf{34.7} & \textbf{49.9} & 28.2 & 13.8 \\
\hline
\end{tabular}
\end{adjustbox}
\caption{Ablation on the self-similarity source used for span generation across three benchmarks, before and after re-ranking.}
\label{tab:ablation_tsm_all}
\end{table*}
\noindent\textbf{\textcolor{black}{Ablation on the self-similarity source for span generation.}}
\textcolor{black}{Tab.~\ref{tab:ablation_tsm_all} compares three variants of the temporal self-similarity map source used for span generation: frame-only~$M^f$, caption-only~$M^c$, and their fusion~$M=\tfrac{1}{2}(M^f+M^c)$.
We report results on the QVHighlights~\cite{lei2021detecting} validation set, Charades-STA~\cite{gao2017tall}, and ActivityNet-Captions~\cite{krishna2017dense}, both before and after the MLLM-based re-ranking stage, while keeping all other components unchanged.
Before re-ranking, the fused setting consistently provides the strongest overall performance, achieving the best mAP on QVHighlights and the best mIoU on both Charades and ActivityNet.
This suggests that visual and caption self-similarity provide complementary temporal cues: frame-only self-similarity yields strong structural boundaries, whereas caption-only self-similarity can capture semantic consistency but is less stable when used alone.
After re-ranking, a similar trend is still observed.
Although frame-only or caption-only sometimes achieves the best score on individual recall metrics, the fused setting still gives the strongest overall and most balanced performance across datasets, especially in mAP/mIoU.
These results support our design choice of constructing the temporal self-similarity map from both frame and caption features, since their fusion yields more robust candidate spans for moment retrieval.}

\subsection{\textcolor{black}{Encoder robustness and ablation studies}}


\begin{table}[t]
\centering
\begin{adjustbox}{width=1.0\textwidth}
\renewcommand{\arraystretch}{1.1}
\begin{tabular}{l c c c c c c}
\hline
\textbf{Backbone} & \textbf{\#Params} & \textbf{Scoring} & \textbf{mAP} & \textbf{mAP@0.5} & \textbf{R1@0.5} & \textbf{R1@0.7} \\
\hline
\multirow{2}{*}{CLIP-B/32~\cite{radford2021clip}} 
& \multirow{2}{*}{87M}
& QMS only  & 31.5 & 49.2 & 42.9 & 29.0 \\
& & QMS + SMS & \textbf{35.3} & \textbf{54.3} & \textbf{54.1} & \textbf{37.9} \\
\hline
\multirow{2}{*}{SigLIP-B/16~\cite{zhai2023sigmoid}}
& \multirow{2}{*}{200M}
& QMS only  & 31.3 & 48.6 & 42.5 & 28.3 \\
& & QMS + SMS & \textbf{34.8} & \textbf{53.8} & \textbf{53.6} & \textbf{36.8} \\
\hline
\multirow{2}{*}{DINOv2-Base~(Ours)~\cite{oquab2023dinov2}}
& \multirow{2}{*}{86M}
& QMS only  & 38.6 & 59.4 & 59.5 & 42.1 \\
& & QMS + SMS & \textbf{38.7} & \textbf{59.6} & \textbf{60.3} & \textbf{42.9} \\
\hline
\end{tabular}
\end{adjustbox}
\caption{Ablation on the visual encoder used to extract frame features for constructing the temporal self-similarity matrix.}
\label{tab:ablation_vis_encoder}
\end{table}
\noindent\textbf{\textcolor{black}{Robustness of the self-matching score across visual encoders.}}
\textcolor{black}{Tab.~\ref{tab:ablation_vis_encoder} evaluates the robustness of Self-SiMS under different visual encoders on the QVHighlights~\cite{lei2021detecting} validation set.
Specifically, we compare QMS alone with the combination of QMS and SMS before the re-ranking stage.}
Our DINOv2-based setting consistently outperforms CLIP and SigLIP, even though the latter use comparable or larger model sizes.
In particular, DINOv2 captures temporal variations more effectively, whereas CLIP and SigLIP tend to produce temporally smoother features~\cite{chatterjee2025streaming}.
Importantly, self-matching span score~(SMS) yields consistent gains across all evaluated encoders, indicating that the benefit of self-similarity-based span scoring is robust to the choice of visual backbone.
Among the tested backbones, DINOv2-Base achieves the best overall performance, demonstrating that it provides the most favorable balance between temporal sensitivity and feature quality in our training-free setting. \\




\begin{table}[t]
\centering
\begin{adjustbox}{width=1.0\textwidth}
\renewcommand{\arraystretch}{1.1}
\begin{tabular}{l l c c c c c}
\hline
\textbf{Method} & \textbf{Text encoder} & \textbf{\#Params} & \textbf{mAP} & \textbf{mAP@0.5} & \textbf{R1@0.5} & \textbf{R1@0.7} \\
\hline
\multirow{3}{*}{VTG-GPT~\cite{Xu2024vtggpt}}
& E5-Mistral~\cite{wang2024improvingtextembeddingslarge} & 7B & 30.87 & 50.80 & 51.94 & 33.16 \\
& NV-Embed~\cite{lee2025nvembed} & 7B & 31.44 & 53.48 & 55.42 & 34.45 \\
& Sentence-BERT~\cite{reimers2019sentence} & 82M & 34.9 & 56.56 & 59.03 & 38.9 \\
\hline
\end{tabular}
\end{adjustbox}
\caption{Ablation on the text encoder used to compute query-caption similarity in the query-dependent VTG-GPT baseline.}
\label{tab:text_encoder}
\vspace{-0.8cm}
\end{table}

\noindent\textbf{\textcolor{black}{Effect of text encoder capacity on the language-style gap.}}
\textcolor{black}{The language-style~(LS) gap in our setting is not merely a surface-level difference in wording, but also arises from the inherent variability of MLLM-generated captions across visually similar frames.
As a result, their shifting focus and granularity make query-caption~(QC) alignment noisy and temporally unstable.
This motivates an additional analysis of whether stronger text encoders alone can stabilize QC alignment.
To isolate this effect, we conduct an additional study on the QVHighlights~\cite{lei2021detecting} validation set using VTG-GPT~\cite{Xu2024vtggpt}, a query-dependent baseline whose span generation and scoring both rely solely on QC similarity.
We adopt VTG-GPT in this analysis because its official implementation is publicly available, whereas the closely related Moment-GPT~\cite{momentGPT} does not provide an official code release.
This setting removes the influence of our self-similarity-based span proposal and scoring, allowing us to directly examine the contribution of the text encoder to QC matching.
As shown in Tab.~\ref{tab:text_encoder}, replacing the default Sentence-BERT encoder with larger alternatives, such as E5-Mistral~\cite{wang2024improvingtextembeddingslarge} and NV-Embed~\cite{lee2025nvembed}, does not improve performance.
These results indicate that the LS gap cannot be sufficiently resolved by stronger text encoding alone.}

\subsection{\textcolor{black}{MLLM ablation studies and Computational efficiency analysis}}
\vspace{-0.4cm}

\begin{table}[!htbp]
\centering
\renewcommand{\arraystretch}{1.2}
\begin{adjustbox}{width=1.0\textwidth}
\begin{tabular}{l l l r r r r}
\hline
\textbf{Setting} & \textbf{Captioning} & \textbf{Re-ranking} & \textbf{mAP} & \textbf{mAP@0.5} & \textbf{R1@0.5} & \textbf{R1@0.7} \\
\hline
Moment-GPT~\cite{momentGPT} & MiniGPT-v2~\cite{minigptv2} & Video-ChatGPT~\cite{video_chatgpt} & 35.9 & 55.7 & 58.9 & 38.6 \\
\hline
Ours & MiniGPT-v2~\cite{minigptv2} & -- & 38.6 & 59.9 & 59.9 & 42.3 \\
 & LLaMA3.2-Vision~\cite{grattafiori2024llama3} & -- & \textbf{38.7} & \textbf{59.6} & \textbf{60.3} & \textbf{42.9} \\
\hline
Ours & MiniGPT-v2~\cite{minigptv2} & MiniGPT-v2~\cite{minigptv2} & \textbf{40.0} & \textbf{62.3} & \textbf{65.0} & \textbf{45.2} \\
 & LLaMA3.2-Vision~\cite{grattafiori2024llama3} & LLaMA3.2-Vision~\cite{grattafiori2024llama3} & {39.3} & {60.5} & {61.0} & {43.2} \\
\hline
\end{tabular}
\end{adjustbox}
\caption{Ablation on MLLM choice for captioning and re-ranking on QVHighlights validation.}
\label{tab:mllm_qvh}
\vspace{-1.6cm}
\end{table}


\begin{table*}[!htbp]
\centering
\renewcommand{\arraystretch}{1.3}
\begin{adjustbox}{width=\textwidth}
\begin{tabular}{l l l | c c c c | c c c c}
\hline
\multirow{2}{*}{\textbf{Setting}} & \multirow{2}{*}{\textbf{Captioning}} & \multirow{2}{*}{\textbf{Re-ranking}} 
& \multicolumn{4}{c|}{\textbf{Charades-STA}} 
& \multicolumn{4}{c}{\textbf{ActivityNet-Captions}} \\
\cline{4-11}
& & 
& {mIoU} & {R1@0.3} & {R1@0.5} & {R1@0.7}
& {mIoU} & {R1@0.3} & {R1@0.5} & {R1@0.7} \\
\hline
Moment-GPT~\cite{momentGPT}
& MiniGPT-v2~\cite{minigptv2}
& Video-ChatGPT~\cite{video_chatgpt}
& 36.5 & 58.2 & 38.4 & 21.6
& 30.8 & 48.1 & 31.1 & 14.9 \\
\hline
\multirow{2}{*}{Ours}
& MiniGPT-v2~\cite{minigptv2}
& --
& 38.1 & 57.0 & 35.3 & \textbf{18.9}
& \textbf{33.8} & \textbf{48.5} & \textbf{27.8} & \textbf{13.7} \\
& LLaMA3.2-Vision~\cite{grattafiori2024llama3}
& --
& \textbf{38.8} & \textbf{58.7} & \textbf{37.2} & 18.7
& 33.7 & 48.5 & 27.1 & 13.4 \\
\hline
\multirow{2}{*}{Ours}
& MiniGPT-v2~\cite{minigptv2}
& MiniGPT-v2~\cite{minigptv2}
& 41.3 & 62.4 & 38.7 & 20.7
& 34.5 & 49.4 & 27.8 & \textbf{13.9} \\
& LLaMA3.2-Vision~\cite{grattafiori2024llama3}
& LLaMA3.2-Vision~\cite{grattafiori2024llama3}
& \textbf{41.9} & \textbf{62.7} & \textbf{39.7} & \textbf{21.0}
& \textbf{34.7} & \textbf{49.9} & \textbf{28.2} & 13.8 \\
\hline
\end{tabular}
\end{adjustbox}
\caption{Ablation on MLLM choice for captioning and re-ranking on Charades-STA and ActivityNet-Captions.}
\label{tab:mllm_ablation_cha_anet}
\vspace{-0.6cm}
\end{table*}

\noindent\textbf{MLLM ablation study.}
To verify that the performance improvement of our framework does not simply arise from the choice of a stronger MLLM, we compare different MLLM backbones within our pipeline. Tab.~\ref{tab:mllm_qvh} compares different MLLM choices for captioning and re-ranking on the QVHighlights~\cite{lei2021detecting} validation set, while Tab.~\ref{tab:mllm_ablation_cha_anet} reports the corresponding comparison on Charades-STA~\cite{gao2017tall} and ActivityNet-Captions~\cite{krishna2017dense}. 
Notably, our framework still achieves strong performance when using MiniGPT-v2~\cite{minigptv2}, the same captioning backbone adopted by prior work Moment-GPT~\cite{momentGPT}, indicating that the improvement does not simply come from switching to a larger MLLM. 
The MiniGPT-v2 variant also provides a controlled efficiency comparison with Moment-GPT, as reported in Tab.~\ref{tab:llm_runtime_comparison} of the main paper. Under the same captioning backbone as Moment-GPT, Self-SiMS uses only one 7B model for both captioning and re-ranking, compared with Moment-GPT's 22B model stack, and achieves slightly lower total MLLM/LLM runtime. Its performance remains comparable across benchmarks and is even slightly higher on QVHighlights, showing that our gain is not simply due to a larger backbone or longer inference.

\subsection{\textcolor{black}{Effect of self-similarity scoring on the language-style gap}}


\begin{table}[!htbp]
\centering
\scriptsize
\setlength{\tabcolsep}{1.8pt}
\renewcommand{\arraystretch}{1.1}
\begin{adjustbox}{width=1.0\textwidth}
\begin{tabular}{l l l c c c c}
\hline
\textbf{Subset} & \textbf{Method} & \textbf{Scoring} & \textbf{mAP} & \textbf{mAP@0.5} & \textbf{R1@0.5} & \textbf{R1@0.7} \\
\hline
\multirow{3}{*}{Full}
& Moment-GPT's~\cite{momentGPT} & Mean & 55.7 & 40.9 & 56.7 & 37.1 \\
& Ours & QMS & 59.5 & 42.1 & 59.4 & 38.6 \\
& Ours & + SMS & \textbf{60.3} & \textbf{42.9} & \textbf{59.6} & \textbf{38.7} \\
\hline
\multirow{3}{*}{\shortstack[l]{Severe LS-gap\\subset}}
& Moment-GPT's~\cite{momentGPT} & Mean & 25.9 & 43.7 & 34.0 & 24.5 \\
& Ours & QMS & 31.3 & 51.8 & 43.4 & 30.7 \\
& Ours & + SMS & \textbf{31.7} & \textbf{52.8} & \textbf{45.3} & \textbf{32.6} \\
\hline
\end{tabular}
\end{adjustbox}
\caption{Performance on the full validation set and the severe Language-style~(LS) gap subset, where query-caption similarity is particularly unstable.}
\label{tab:ls_gap_subset}
\vspace{-0.6cm}
\end{table}
\textcolor{black}{To further analyze the effect of self-similarity scoring modeling under a language-style~(LS) gap, we evaluate a subset of queries on the QVHighlights~\cite{lei2021detecting} validation set for which query-caption~(QC) similarity is particularly unstable, corresponding to cases with a severe language-style gap.
We use the first-order-difference energy to measure how sharply the QC score changes over time; a high value indicates that the similarity signal fluctuates strongly even within the target moment.
We define the severe LS-gap subset as queries whose QC similarity is both noisy and weak within the ground-truth span.
Concretely, we select queries whose QC similarity has~(i) first-order-difference energy in the top quartile and~(ii) mean similarity in the bottom quartile.
As shown in Tab.~\ref{tab:ls_gap_subset}, replacing mean scoring scheme used in Moment-GPT~\cite{momentGPT} with query-matching span score~(QMS) consistently improves performance on both the severe LS-gap subset and the full dataset, confirming that QMS provides a more stable span score under unstable QC alignment.
Additionally, adding self-matching span score~(SMS) produces consistent gains in both settings, with the improvement being more pronounced on the severe LS-gap subset, particularly in terms of R1@0.5 and R1@0.7.
These results indicate that self-similarity-based modeling effectively complements query-dependent scoring by stabilizing span-level decisions in hard cases while still providing additional benefits in the general setting. \\}

\subsection{\textcolor{black}{Comparison of span generation strategies}}

\begin{table}[!htbp]
\centering
\begin{tabular}{l c c c c}
\hline
Method & mAP@avg & mAP@0.5 & R1@0.5 & R1@0.7 \\
\hline
Sliding Window~\cite{ren2024timechat} & 12.5 & 30.5 & 29.9 & 10.5 \\
K-means clustering~\cite{nam2021zero} & 31.3 & 47.7 & 49.1 & 32.1 \\

Ours & \textbf{39.3} & \textbf{60.5} & \textbf{61.0} & \textbf{43.2} \\
\hline
\end{tabular}
\caption{Comparison of candidate span generation strategies.}
\label{tab:span_gen_ablation}
\vspace{-0.6cm}
\end{table}

Table~\ref{tab:span_gen_ablation} compares our self-similarity-based span generation strategy against two generic query-agnostic alternatives: sliding-window proposals and clustering-based proposals on the QVHighlights~\cite{lei2021detecting} validation set.
\textcolor{black}{For the sliding-window baseline, we adopt a length-adaptive design inspired by TimeChat~\cite{ren2024timechat}; unlike the fixed temporal scale used in that training-based framework, we set the window size and stride proportionally to the video length as $\lfloor L/3 \rfloor$ and $\lfloor L/6 \rfloor$, respectively, in our training-free setting.
For the clustering-based baseline, we follow PSVL~\cite{nam2021zero} and set the number of clusters to~$k=5$.
}
Sliding windows serve as a simple baseline that uniformly covers the temporal axis with fixed candidate spans, while clustering-based proposals group temporally similar features without explicitly modeling boundary transitions.
In contrast, our method generates candidate spans from self-similarity-derived boundary scores, producing more precise temporal proposals and leading to best moment retrieval performance.
\end{document}